\documentclass[lettersize,journal]{IEEEtran}
\usepackage{amsmath,amsfonts}
\usepackage{algorithmic}
\usepackage{algorithm}
\usepackage{array}
\usepackage[caption=false,font=normalsize,labelfont=sf,textfont=sf]{subfig}
\usepackage{textcomp}
\usepackage{stfloats}
\usepackage{url}
\usepackage{verbatim}
\usepackage{graphicx}
\usepackage{cite}
\usepackage{multirow}
\usepackage{makecell}
\usepackage{arydshln}
\usepackage{subcaption}
\expandafter\def\csname ver@subfig.sty\endcsname{}
\usepackage{array}
\usepackage{subfig}
\usepackage{caption}
\usepackage{subcaption}
\usepackage{subfig}
\usepackage{wrapfig}
\usepackage{graphicx}
\usepackage[colorlinks=true, linkcolor=blue, citecolor=blue, urlcolor=blue]{hyperref}
\usepackage{xcolor}
\usepackage{pifont}

\newcommand{\eg}{\textit{e.g.}}
\newcommand{\ie}{\textit{i.e.}}
\renewcommand{\paragraph}[1]{\vspace{1.0ex}\noindent\textbf{#1.}}

\begin{document}

\title{A Simple Baseline with Single-encoder for Referring Image Segmentation}

\author{Seonghoon Yu, Ilchae Jung, Byeongju Han, Taeoh Kim, Yunho Kim, \\ Dongyoon Wee, and Jeany Son
\thanks{S. Yu, and J. Son are with AI Graduate School, GIST, South Korea (email: seonghoon@gm.gist.ac.kr; jeany@gist.ac.kr).}
\thanks{I. Jung, B. Han, T. Kim, and D. Wee are with NAVER Cloud, South Korea (email: ilchae.jung@navercorp.com; byeongju.han@navercorp.com; taeoh.kim@navercorp.com; dongyoon.wee@navercorp.com).}
\thanks{Y. Kim is with Electrical Engineering and Computer Science, GIST, South Korea (email: youknowyunho@gm.gist.ac.kr).}
\thanks{This work is done when S. Yu is an intern at NAVER Cloud.}
}



\markboth{Journal of \LaTeX\ Class Files,~Vol.~\#\#, No.~\#, August~2024}%
{Shell \MakeLowercase{\textit{et al.}}: A Sample Article Using IEEEtran.cls for IEEE Journals}


\maketitle
\IEEEpubid{
   \parbox{\textwidth}{\centering
   \vspace{0.2cm}
   0000--0000~\copyright~2024 IEEE. This work has been submitted to the IEEE for possible publication.\\
   Copyright may be transferred without notice, after which this version may no longer be accessible.}
}
\IEEEpubidadjcol
\begin{abstract}
Referring image segmentation (RIS) requires dense vision-language interactions between visual pixels and textual words to segment objects based on a given description.
However, commonly adapted dual-encoders in RIS, \eg~Swin transformer and BERT (uni-modal encoders) or CLIP (a multi-modal dual-encoder), 
lack dense multi-modal interactions during pre-training, leading to a gap with a pixel-level RIS task.
To bridge this gap, existing RIS methods often rely on multi-modal fusion modules that interact two encoders, but this approach leads to high computational costs. 
In this paper, we present a novel RIS method with a single-encoder, \ie~BEiT-3, maximizing the potential of shared self-attention across all framework components.
This enables seamless interactions of two modalities from input to final prediction, producing granularly aligned multi-modal features.
Furthermore, we propose lightweight yet effective decoder modules, a Shared FPN and a Shared Mask Decoder, which contribute to the high efficiency of our model.
Our simple baseline with a single encoder achieves outstanding performances on the RIS benchmark datasets while maintaining computational efficiency, compared to the most recent SoTA methods based on dual-encoders.

\begin{IEEEkeywords}
Referring image segmentation, vision-language, multi-modal interaction.
\end{IEEEkeywords}

\end{abstract}



\section{Introduction}
\label{sec:intro}

\IEEEPARstart{T}{he} goal of Referring Image Segmentation (RIS) is to precisely identify and segment an object within an image based on a specific textual description provided.
Unlike conventional segmentation tasks, RIS uniquely processes free-form, arbitrary-length text, and open-vocabulary to localize only the referenced target object, requiring dense alignment between input words and visual pixels.
This language-driven interaction with visual content is highly applicable in fields such as human-object interaction~\cite{interaction} and image editing~\cite{image_editing}.

\begin{figure}[t]
    \centering
    \includegraphics[width=0.9\linewidth]{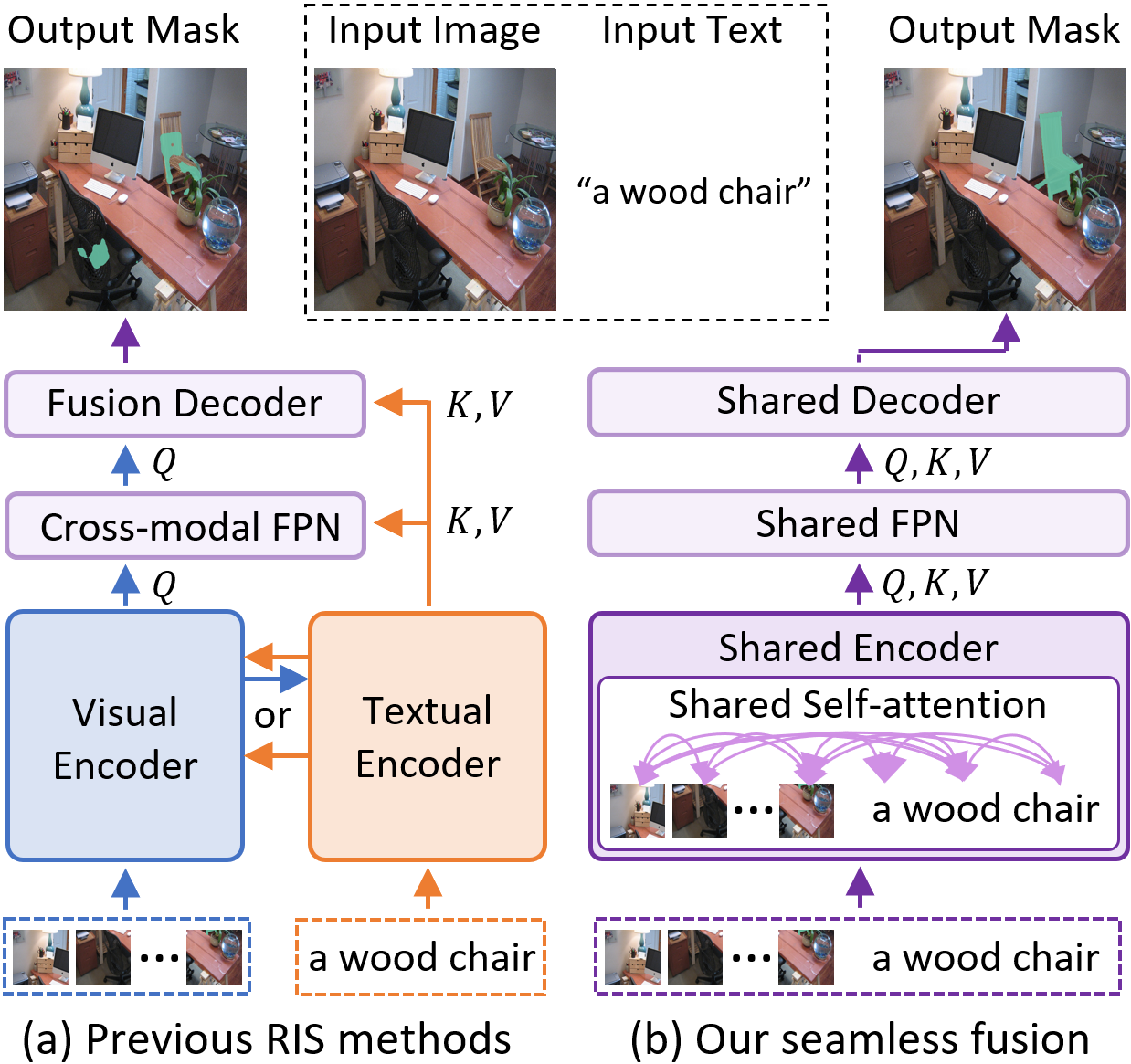}
    \caption{Comparison of previous RIS methods~\cite{joint_4_risclip, early_3_dmmi} and our approach.
    (a) previous methods focus on bridging two encoders with high-cost fusion modules to compensate for coarse interaction at dual-backbones pre-training, resulting in an over-parameterized architecture.
    In contrast, our approach in (b) provides a cost-efficient architecture by fully leveraging shared self-attention across all processes, eliminating the need for interacting two encoders and offering granular alignments at patch-word levels.
    }
    \label{fig:intro_framework}
    \vspace{-0.3cm} 
\end{figure}

\begin{figure*}[t]
     \centering
     \hspace{0.2cm}
     \begin{minipage}[t]{0.31\textwidth}
         \centering
         \includegraphics[width=\textwidth]{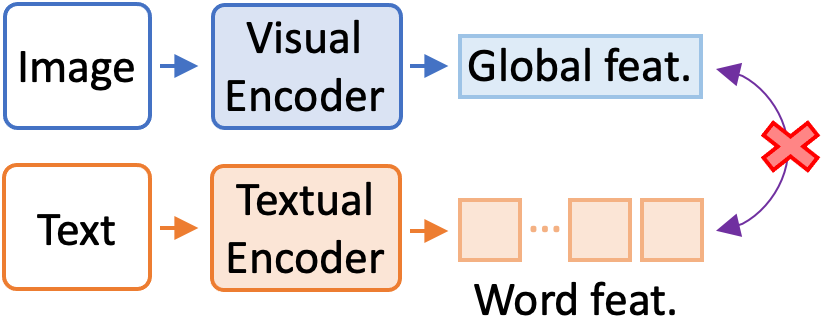}
         \caption*{(a) No interaction}
     \end{minipage}
     \hfill
     \begin{minipage}[t]{0.30\textwidth}
         \centering
         \includegraphics[width=\textwidth]{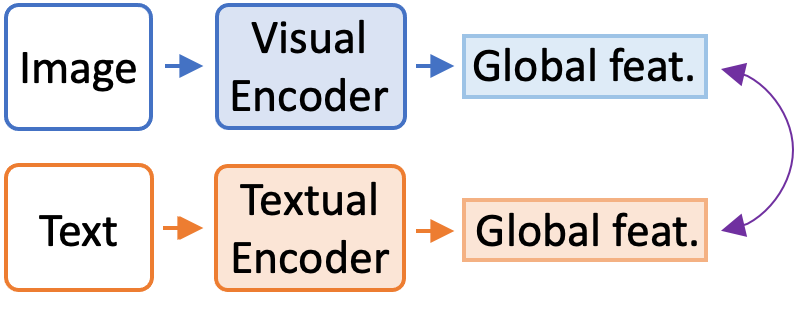}
         \caption*{(b) Global-level interaction}
     \end{minipage}
     \hfill
     \begin{minipage}[t]{0.29\textwidth}
         \centering
         \includegraphics[width=\textwidth]{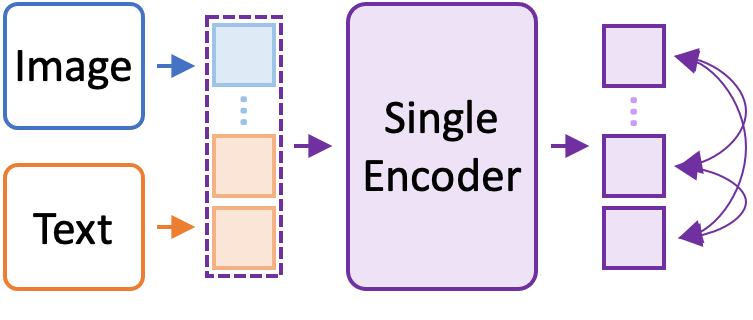}
         \caption*{(c) Dense-level interaction}
     \end{minipage}
     \hfill
        \caption{Illustration of multi-modal interaction levels in a backbone pre-training used in RIS: (a) Two encoders trained independently on each modality with no interaction (\eg, Swin transformer and BERT), (b) The pre-trained model allowing only one global-level interaction between two modalities (\eg, CLIP). Both (a) and (b) lack fine-grained interactions, leading to misalignment with demands of a RIS task, which requires dense interactions at a pixel-word level. Therefore, we propose to utilize a single-encoder in (c) as a backbone for RIS.}
        \label{fig:motivation}
        \vspace{-0.3cm} 
\end{figure*}

Current RIS methods are commonly rooted in \textbf{dual-encoder paradigms}, using separate encoders to extract and fuse different modalities.
However, dual-backbones typically used in RIS are sub-optimal for a RIS task due to the mismatch in a targeted vision-language interaction level between backbone pre-training and the RIS task, as depicted in Fig.~\ref{fig:motivation}.
For example, uni-modal encoders,  \eg, Swin transformer~\cite{swin} and BERT~\cite{bert}, are trained independently without interactions (Fig.~\ref{fig:motivation}a),
and a multi-modal dual-encoder CLIP~\cite{clip} allows only one interaction at the global level for image-sentence alignment (Fig.~\ref{fig:motivation}b).
In contrast, the RIS task demands dense and granular vision-language interactions for pixel-word level alignment~\cite{joint_2_coupalign}.

To exploit dual-backbones with insufficient interactions at pre-training for the RIS task, prior methods~\cite{early_3_dmmi, joint_4_risclip} focus on densely connecting two encoders using cross-attention modules, as shown in Fig.~\ref{fig:intro_framework}a.
However, these approaches result in an over-parameterized architecture with high computational costs due to the substantial complexity of their fusion modules.
Furthermore, despite the success of shared self-attention in vision-language pre-training tasks~\cite{vilt,vlmo,beit3}, which simultaneously encodes and fuses two modalities to learn detailed multi-modal features, this approach has not been explored in RIS. 
\IEEEpubidadjcol
We believe that this shared-fusion method is particularly well-suited for RIS due to its ability to align two modalities at the patch-word level, as required by RIS.

In this paper, we introduce a novel RIS framework that fully leverages shared self-attention from the encoder to the decoder, as depicted in Fig.~\ref{fig:intro_framework}b.
This enables \textbf{seamless interactions} at the granular level of visual patches and textual words across all the framework, ensuring accurate alignment.
By employing a \textbf{single encoder}, namely BEiT-3~\cite{beit3}, that is pre-trained on large-scale datasets with detailed multi-modal interaction via shared self-attention, our framework aligns closely with the RIS task's requirements, as illustrated in Fig.~\ref{fig:motivation}c, significantly enhancing RIS performance.
Moreover, the effectiveness of shared self-attention in our fusion decoder, Shared FPN and Shared Mask Decoder modules, contributes to the \textbf{lightweight} nature of our final model.

Our simple baseline with a single-encoder not only delivers superior performance on the RIS benchmark datasets by borrowing the capability of shared self-attention but also exhibits high efficiency in both parameters and computational cost, thanks to our lightweight FPN and decoder, compared to SoTA dual-encoder based methods.

Our contributions can be summarized as follows:
\begin{itemize}
    \item We propose a novel \textit{single-encoder paradigm} for RIS, aimed at resolving the discrepancy in a desired level of multi-modal interactions between a pre-training of backbones and requirements of a RIS task, which exists in current dual-encoder frameworks.

    \item 
    {
    We introduce \textit{seamless fusion} approach that continuously interacts two modalities from input to final prediction, ensuring consistently well-aligned features at the patch-word level across all processes.
    }

    \item 
    Our \textit{lightweight} decoders, Shared FPN and Shared Mask Decoder, {that exploit shared self-attention}, have minimal computational overhead while achieving high performance, validating that shared self-attention is an advantageous fusion method for RIS. 

    \item Our framework, featured by \textit{simple yet effective design}, achieves superior performance across various RIS benchmark datasets and exhibits high efficiency in computational demands (\ie, parameters and FLOPs).

\end{itemize}

\section{Related Work}
\label{sec:related}

\subsection{Referring Image Segmentation}


Supervised RIS approaches~\cite{fully_ris_5, fully_ris_6, fully_ris_7, fully_ris_8, 80_classes, pseudo-ris} focus on fusing two modality features into a RIS framework based on dual-encoders.
The pioneering works~\cite{pior_ris_1, pior_ris_2} simply concatenate two modalities features and forward them into a segmentation head~\cite{seg_head}.
CRIS~\cite{late_3_cris} propose a CLIP-driven framework, leveraging vision-language aligned features via contrastive learning~\cite{clip}.
CGFormer~\cite{late_9_cgformer} captures object-relations with given textual semantics via grouping strategies.
LAVT~\cite{early_1_lavt} introduces a first early fusion approach, where the text features are integrated into each stage of a visual encoder.
DMMI~\cite{early_3_dmmi} designs dual cross-attention modules enabling the information flow in two directions.
CoupAlign~\cite{joint_2_coupalign} aims to achieve multi-level alignments by allowing multi-modal interactions at each stage of two encoders, jointly.
RISCLIP~\cite{joint_4_risclip} presents CLIP-based methods that jointly interact with two encoders.
Unlike these prior works, our methods tackle the mismatch in a desired multi-modal interaction level between dual-backbone pre-training and the requirements of RIS, and aim to provide denser and seamless interactions of two modalities and offer an efficient model.

\subsection{Shared Self-attention}
Shared self-attention~\cite{vilt} is a simple fusion approach of multi-modalities.
This simultaneously encodes and merges two modalities by passing concatenated modalities into a self-attention layer.
By using this, several works~\cite{vilt, vlmo, beit3} in vision-language pre-training tasks achieve deeper multi-modal interactions within a single architecture, enhancing the learning of fine-grained multi-modal representations.
Recent works~\cite{vlmo, beit3} combine the shared fusion with mixture of experts~\cite{moe, switch_transformer} to capture specific modality information, showing a remarkable ability to learn detailed representations.
However, most RIS works still rely on cross-attention modules such as transformer decoder layers for multi-modal interactions.
In contrast, our work first explores the effectiveness of shared self-attentions in a vision-language dense prediction task like RIS by integrating it into all components of our framework.

\section{Method}
\label{sec:method}

\begin{figure*}[t]
    \centering
    \includegraphics[width=0.92\linewidth]{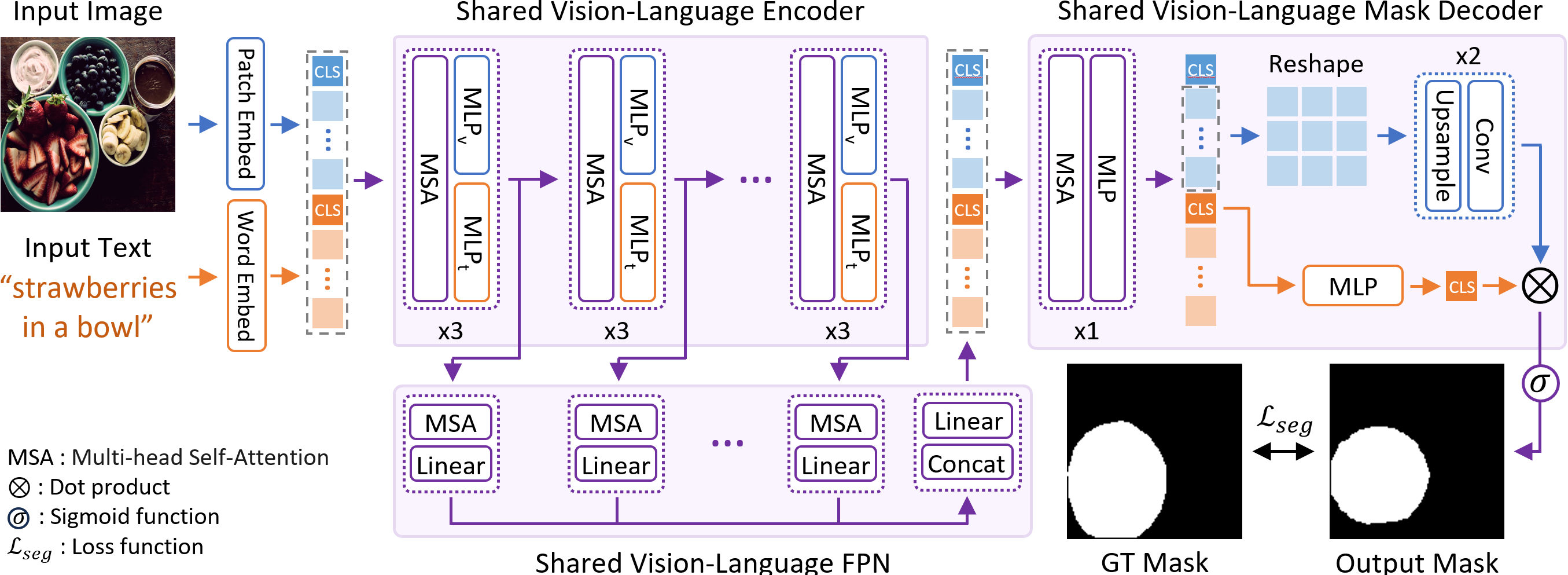}
    \caption{Overall framework of our Shared-RIS. Given an input image and text, we embed and concatenate two modality inputs, and pass a concatenated embedding to a shared vision-language encoder to extract shared visual-textual features. We further refine these features at multiple levels using a shared vision-language FPN, and then we generate a segmentation mask using our shared vision-language mask decoder. Note that, our framework has seamless multi-modal fusions across all its components.}
    \label{fig:framework}
    \vspace{-0.3cm}
\end{figure*}

In this section, we introduce Shared-RIS, the first framework with a single-encoder for referring image segmentation (RIS) that deviates from conventional dual-encoder paradigms in RIS.
At the core of Shared-RIS lies a \textit{shared transformer} architecture, which processes concatenated multi-modal inputs or features.
This design maximizes the potential of shared-fusion in aligning and understanding two modalities.
As shown in Fig.~\ref{fig:framework}, our framework consists of three components to generate segmentation masks grounded to a given expression: 1) \textit{Shared Vision-Language Encoder} (Sec.~\ref{subsec:shared_encoder}), 2) \textit{Shared FPN} (Sec.~\ref{subsec:shared_fpn}), and 3) \textit{Shared Mask Decoder} (Sec.~\ref{subsec:shared_decoder}).
These are then trained using a loss function (Sec.~\ref{subsec:loss_function}). 


A major challenge in current RIS approaches is obtaining fine-grained alignments between pre-trained vision and text encoders.
This issue has been tackled through various fusion strategies, including early fusions~\cite{early_1_lavt, early_3_dmmi}, and joint fusions~\cite{joint_2_coupalign, joint_3_crossvlt} (for more details on these strategies, please refer to Fig.~\ref{fig:fusion_categories}).
However, due to computational overhead, these methods typically involve only a few intermediate fusions via cross-attention modules between separate encoders.

In contrast, our framework aims to provide a more dense fusion strategy, termed \textit{seamless fusion}, while maintaining minimal complexity.
Our method simultaneously encodes and merges two modalities across all framework components from the first input to the final prediction, employing shared self-attention techniques~\cite{vilt}.
We believe that the property of shared self-attention, which allows granular token-level interactions between two modalities, is promising for RIS. 
In the following section, we further elaborate on each component in our framework in detail.

\subsection{Shared Vision-Language Encoder}
\label{subsec:shared_encoder}
We first extract shared vision-language features from a given multi-modal set via a shared vision-language encoder, namely BEiT-3~\cite{beit3}.
The use of this single encoder offers several advantages: (1) During pre-training of the backbone, it is trained with the same level of interactions required for RIS at scale, improving pixel-word level alignment.
(2) During fine-tuning on RIS, it allows dense vision-language interactions at token levels across all layers, thanks to two modalities being encoded in a shared manner with shared self-attention. 
These two advantages allow the extraction of valuable and fine-grained visual-textual features, which is highly beneficial for RIS, overcoming the limited size of RIS datasets.
In this section, we demonstrate the process of extracting these shared visual-textual features using a shared vision-language encoder.

\paragraph{Visual Patch Embedding}
First, the input image $I \in \mathbb{R}^{h \times w \times 3}$ is reshaped into a sequence of flattened image patches $\mathbf{x} \in \mathbb{R}^{n \times (p^2 \times 3)}$, where $p$ denotes a patch size, and $n=hw/p^2$ is the number of patches. All patches are linearly projected using a projection layer $\mathbf{P} \in \mathbb{R}^{(p^2 \times 3) \times d}$, where $d$ is a channel dimension, and
a visual class token $\mathbf{v}_{cls} \in \mathbb{R}^{1 \times d} $ is prepended.
Visual patch embeddings $\mathbf{v}_{0} \in \mathbb{R}^{(n+1) \times d}$ are then obtained by adding learnable visual position embeddings $\mathbf{v}_{pos} \in \mathbb{R}^{(n+1) \times d}$ as follows:
\begin{equation}
    \mathbf{v}_{0} = \left[\mathbf{v}_{cls}, \mathbf{x}_1\mathbf{P}, \dots, \mathbf{x}_n\mathbf{P} \right] + \mathbf{v}_{pos}.
\end{equation}

\paragraph{Textual Token Embedding}
The input text $\mathbf{y} \in \mathbb{R}^{m}$ is tokenized using SentencePiece~\cite{sentencepiece},  where $m$ is the length of textual input.
Then we frame the tokenized text sequence with a textual class token $\mathbf{t_{cls}} \in \mathbb{R}^{1 \times d}$ at the start and an end-of-token $\textbf{t}_{eot} \in \mathbb{R}^{1 \times d}$ at the end.
Textual token embeddings $\mathbf{t}_{0} \in \mathbb{R}^{(m+2) \times d}$ are then derived by incorporating learnable textual position embeddings $\mathbf{t}_{pos} \in \mathbb{R}^{(m + 2) \times d}$ by adding similar to visual patch embedding, as follows:
\begin{equation}
    \mathbf{t}_{0} = \left[\mathbf{t}_{cls}, \mathbf{t}_1, \dots, \mathbf{t}_m, \mathbf{t}_{eot} \right] + \mathbf{t}_{pos}.
\end{equation}

\paragraph{Visual-Textual Embedding}
We concatenate the visual patch embedding and the textual token embedding to obtain a single visual-textual embedding $\mathbf{h}_0$ as follows:
\begin{equation}
    \mathbf{h}_{0} = \left[\mathbf{v}_{0} ; \  \mathbf{t}_{0} \right] \in \mathbb{R}^{(n+m+3)\times d},
\end{equation}
where $\left[ \cdot ; \cdot \right]$ denotes the operation of concatenation.

\paragraph{Shared Transformer}
Finally, we forward visual-textual embeddings $\mathbf{h}_{0}$ into a sequence of $L$ shared transformers, \ie, multiway transformers~\cite{beit3}, to extract shared visual-textual features $\mathbf{h}_{i} \in \mathbb{R}^{(n+m+3) \times d}$, $i \in \{1, \dots, L \}$:
\begin{align}
    &\mathbf{h}^{\prime}_i = \text{MSA}(\text{LN}(\mathbf{h}_{i-1})) + \mathbf{h}_{i-1},\\
    &\mathbf{h}_{i} = \left[\text{MLP}_v(\text{LN}(\mathbf{v}^{\prime}_i)) ; \  \text{MLP}_t(\text{LN}(\mathbf{t}^{\prime}_i))  \right] +  \mathbf{h}^{\prime}_i, \label{eq:two_mlp}
\end{align}
where $\mathbf{v}^{\prime}_i \in \mathbb{R}^{(n+1)\times d}$ and $\mathbf{t}^{\prime}_i \in \mathbb{R}^{(m+2)\times d}$ is obtained by splitting $\mathbf{h}^{\prime}_i \in \mathbb{R}^{(n+m+3)\times d}$ into two modalities.
MSA$(\cdot)$ denotes multi-head self-attention, sharing two modalities, LN$(\cdot)$ is layer normalization, and MLP$_v$$(\cdot)$ and MLP$_t$$(\cdot)$ are two separate multi-layer perceptrons for each modality.

\subsection{Shared Vision-Language FPN}
\label{subsec:shared_fpn}
In this section, we introduce a shared vision-language FPN for referring image segmentation, uniquely designed with shared self-attention modules, as shown in Fig.~\ref{fig:framework}.
Unlike previous FPN methods in RIS~\cite{late_3_cris, early_1_lavt} which inject textual features into multi-level visual features using cross-attention, our method seamlessly incorporates both visual and textual features at multiple levels with a single self-attention layer, showing that even a simple shared layer alone is highly effective in learning and capturing detailed visual-semantic representations for RIS.

We first extract shared visual-textual features $\mathbf{h}_j$ from outputs of intermediate stages $j$ in a single shared encoder described in Section~\ref{subsec:shared_encoder}.
Each stage feature is then projected through their respective linear layer $\mathbf{W}_j \in \mathbb{R}^{d \times d_w}$, to reduce the dimension from $d$ to $d_w$, and followed by layer normalization and multi-head self-attention.
Subsequently, all outputs are concatenated along the channel dimension and passed to a final linear layer $\mathbf{W}_f \in \mathbb{R}^{(K \times d_w) \times d_w}$, which changes dimensions of the concatenated feature from $K \times d_w$ to $d_w$:
\begin{align}
    &\mathbf{z}_j = \text{MSA}(\text{LN}(\mathbf{h}_j\mathbf{W}_j)) + \mathbf{h}_j\mathbf{W}_j, \\ 
    &\mathbf{z} =  \left[ \mathbf{z}_{l_1}^T ; \ \dots ; \ \mathbf{z}_{l_K}^T \right]^T\mathbf{W}_{f}, 
\end{align}
where $j$ denotes the intermediate stage indices of the encoder, $j \in \{l_1, \dots, l_K\}$, $K$ is a pre-defined number of intermediate stages, and $\mathbf{z} \in \mathbb{R}^{(n+m+3) \times d_w}$ is aggregated shared visual-textual features at multiple levels.


\subsection{Shared Vision-Language Mask Decoder}
\label{subsec:shared_decoder}
We introduce a shared vision-language mask decoder that is designed with shared self-attention, to decode $\mathbf{z}$ into segmentation masks, as illustrated in Fig.~\ref{fig:framework}.
Prior studies~\cite{early_3_dmmi, late_9_cgformer} fuse different modalities in their fusion decoder with multiple cross-attention modules (\ie, transformer decoder layers), which increases computational costs.
In contrast, we exploit only one shared fusion layer within the mask decoder and again validate its effectiveness in learning fine-grained features at patch-word levels.




\paragraph{Shared Transformer}
In our mask decoder, we employ only one shared transformer layer sharing two modalities with one shared MLP, instead of two separate MLPs in a shared encoder:
\begin{align}
    &\mathbf{u}^{\prime} = \text{MSA}(\text{LN}(\mathbf{z})) + \mathbf{z}, \\
    &\mathbf{u} = \text{MLP}(\text{LN}(\mathbf{u}^{\prime})) + \mathbf{u}^{\prime},
\end{align}
where $\mathbf{u} \in \mathbb{R}^{(n+m+3) \times d_w}$ is a refined shared visual-textual feature.
It is split into two modality features as visual feature $\mathbf{v}_{u} \in \mathbb{R}^{(n+1) \times d_w}$ and textual feature $\mathbf{t}_{u} \in \mathbb{R}^{(m+2) \times d_w}$ to utilize them in the next mask prediction.

\paragraph{Mask Prediction}
To segment the regions described by expressions, we obtain the visual feature map $\mathbf{F}_0 \in \mathbb{R}^{(h/p) \times (w/p) \times d_w}$ by excluding the visual class token from $\mathbf{v}_u$ and reshaping the remaining $n$ visual feature tokens into a 2D feature map. The feature map is scaled up using an upsampling layer (Upsample) and a convolutional block (Conv), which includes a convolutional layer with ReLU activation and batch normalization:
\begin{align}
    &\mathbf{F}_i = \text{Conv}_i(\text{Upsample}_i(\mathbf{F}_{i-1})), ~~ i = 1, \dots l ,
\end{align}
where $l$ is set to 2. 
Finally, we predict the probability map $M$ that highlights the referred regions by computing similarity between the feature at each pixel in $\mathbf{F}_l$ and the textual class token $\mathbf{t}^{cls}_{u}$. 
In this paper, we further update $\mathbf{t}^{cls}_{u}$ using MLP, \textit{i.e.} $\mathbf{t}'^{cls}_{u} = \text{MLP}(\mathbf{t}^{cls}_{u})$, and compute the similarities by dot product between features, as follows:
\begin{align}
    &M(i,j) = \sigma(\mathbf{F}_l(i,j) \otimes \mathbf{t}'^{cls}_{u}),
\end{align}
where $\otimes$ and $\sigma(\cdot)$ denote dot-product and a sigmoid function, respectively. 
During inference, the binary segmentation mask is generated by applying a threshold $\tau$ to the probability map $M$ and then resizing it to align with the dimensions of the input image ($h, w$).

\subsection{Loss Function}
\label{subsec:loss_function}
In a referring image segmentation task, each pixel in an image is categorized as either the background or the foreground, \ie, a pixel-level binary classification.
To address this, our approach utilizes a linear combination of binary cross-entropy loss $\mathcal{L}_{\text{bce}}$ and a dice loss~\cite{dice_loss} $\mathcal{L}_{\text{dice}}$ between a predicted probability map $M$ and a corresponding ground truth mask $\hat{M}$:
\begin{align}
    &\mathcal{L}_{\text{seg}} = \lambda_{\text{bce}} \mathcal{L}_{\text{bce}}(M, \hat{M}) + \lambda_{\text{dice}} \mathcal{L}_{\text{dice}}(M, \hat{M}),
\end{align}
where $\lambda_{\text{bce}}$ and $\lambda_{\text{dice}}$ are hyperparameters balancing two losses.

\section{Experiments}
\label{sec:experiments}



\begin{table*}[t]
    \centering
    \small
    \caption{oIoU comparison with other RIS methods on RefCOCO, RefCOCO+, and RefCOCOg datasets. U and G indicate UMD and Google partitions of RefCOCOg. The best performances are in \textbf{bold} and the second best are \underline{underlined}.
    Explanations of each fusion category are detailed and illustrated in Fig.~\ref{fig:fusion_categories}.
    The mIoU results are also reported in our supplementary.
    }
    \scalebox{0.995}{
    \begin{tabular}{l|l|l|ccc|ccc|ccc}
    \hline
    \multirow{2}{*}{Methods}  & \multicolumn{2}{c|}{Encoders}& \multicolumn{3}{c|}{RefCOCO}& \multicolumn{3}{c|}{RefCOCO+} & \multicolumn{3}{c}{RefCOCOg} \\ \cline{2-12}
         & Visual & Textual & val & testA & testB & val & testA & testB & val(U) & test(U) & val(G) \\ \hline
    \multicolumn{12}{l}{\textit{Late Fusions}} \\ 
    \cdashline{1-12}[0.2pt/1pt]
    ReSTR~\cite{late_2_restr} & ViT-B & BERT & 67.22 & 69.30 & 64.45 & 55.78 & 60.44 & 48.27 & - & - & 54.48 \\
    %
    CRIS~\cite{late_3_cris} & CLIP \tiny R101 & CLIP & 67.35 & 71.54 & 62.16 & 57.94 & 64.05 & 48.42 & 56.56 & 57.38 & - \\
    BKINet~\cite{late_5_bkinet} & CLIP \tiny R101 & CLIP & 73.22 & 76.43 & 69.42 & 64.91 & 69.88 & 53.39 & 64.21 & 63.77 & 61.64 \\
    CM-MaskSD~\cite{late_4_cmmasksd} & CLIP \tiny ViT-B & CLIP & 72.18 & 75.21 & 67.91 & 64.47 & 69.29 & 56.55 & 62.47 & 62.69 & - \\
    VLT~\cite{late_7_vlt} & Swin-B & BERT & 72.96 & 75.96 & 69.60 & 63.53 & 68.43  & 56.92 & 63.49 & \underline{66.22} & \underline{62.80} \\
    PCAN~\cite{late_8_pcan} & Swin-B & BERT & 73.71 & 76.26 & 70.47 & 64.01 & 70.01 & 54.81 & 64.43 & 65.68 & 62.76 \\
    ReLA~\cite{late_1_ReLA} & Swin-B & BERT & 73.92 & 76.48 & 70.18 & \underline{66.04} & \underline{71.02} & \underline{57.65} & \underline{65.00} & 65.97 & 62.70 \\
    CGFormer~\cite{late_9_cgformer} & Swin-B & BERT & \underline{74.75} & \underline{77.30} & \underline{70.64} & 64.54 & 71.00 & 57.14 & 64.68 & 64.09 & 62.51 \\ 
    \hline
    \multicolumn{12}{l}{\textit{Early Fusions}} \\ 
    \cdashline{1-12}[0.2pt/1pt]
    LAVT~\cite{early_1_lavt} & Swin-B & BERT & 72.73 & 75.82 & 68.79 & 62.14 & 68.38 & 55.10 & 61.24 & 62.09 & 60.50 \\
    SLViT~\cite{early_2_slvit} & SegNeXt-B & BERT & 74.02 & 76.91 & 70.62 & 64.07 & 69.28 & 56.14 & 62.75 & 63.57 & 60.94 \\
    SADLR~\cite{early_4_sadlr} & Swin-B & BERT & 74.24 & 76.25 & 70.06 & 62.48 & 69.09 & 55.19 & 63.60 & 63.56 & 61.16 \\ 
    DMMI~\cite{early_3_dmmi} & Swin-B & BERT & 74.13 & 77.13 & 70.16 & 63.98 & 69.73 & 57.03 & 63.46 & 64.19 & 61.98 \\ 
    \hline
    \multicolumn{12}{l}{\textit{Joint Fusions}} \\ 
    \cdashline{1-12}[0.2pt/1pt]
    ETRIS~\cite{joint_1_etris} & CLIP \tiny R101 & CLIP & 69.36 & 72.68 & 63.68 & 58.59 & 66.93 & 50.14 & 58.62 & 60.12 & 55.56 \\
    CrossVLT~\cite{joint_3_crossvlt} & Swin-B & BERT & 73.44 & 76.16 & 70.15 & 63.60 & 69.10 & 55.23 & 62.68 & 63.75 & - \\
    CoupAlign~\cite{joint_2_coupalign} & Swin-B & BERT & 74.70 & \textbf{77.76} & 70.58 & 62.92 & 68.34 & 56.69 & 62.84 & 62.22 & - \\
    RISCLIP~\cite{joint_4_risclip} & CLIP \tiny ViT-B & CLIP & 73.57 & 76.46 & 69.76 & 65.53 & 70.61 & 55.49 & 64.10 & 65.09 & - \\
    \hline
    \multicolumn{12}{l}{\textit{Seamless Fusion}} \\ \cdashline{1-12}[0.2pt/1pt]
    Shared-RIS (Ours)  & \multicolumn{2}{c|}{BEiT-3 \tiny ViT-B} & \textbf{74.83} & 76.83 & \textbf{71.83} & \textbf{68.42} & \textbf{72.83} & \textbf{61.63} & \textbf{68.66} & \textbf{69.38} & \textbf{65.65} \\
    \hline
    \end{tabular}
    }
\label{tab:main_results}
    \vspace{-0.1cm}
\end{table*}

\begin{table*}[t]
    \centering
    \small
    \caption{oIoU comparison with the RIS methods leveraging additional pre-trained models (\eg, LLM or SAM).
    $\dagger$ denotes LLM-based RIS models trained on extra segmentation data such as ADE20k and COCO-Stuff in addition to the combined RefCOCO datasets. In contrast, our methods are trained and evaluated on each individual RefCOCO dataset.
    $\ddagger$ indicates the performance when our methods are trained on the combined RefCOCO datasets.
    * denotes public codes are not available.
    }
    \scalebox{0.995}{
    \begin{tabular}{l|c|c|ccc|ccc|ccc}
    \hline
    \multirow{2}{*}{Methods}  & \multirow{2}{*}{Encoders} & \multirow{2}{*}{GFLOPs} & \multicolumn{3}{c|}{RefCOCO}& \multicolumn{3}{c|}{RefCOCO+} & \multicolumn{3}{c}{RefCOCOg} \\ \cline{4-12}
         & & & val & testA & testB & val & testA & testB & val(U) & test(U) & test(G) \\ \hline
    LISA-7B$^\dagger$~\cite{lisa}  &  CLIP \text{\tiny ViT-L} + SAM  & 13,026 & 74.1  & 76.5 &  71.1 & 62.4 & 67.4 & 56.5 & 66.4 & 68.5 & -  \\
    GSVA-7B$^{\dagger *}$~\cite{gsva}  & CLIP \text{\tiny ViT-L} + SAM & - & 76.4  & 77.4 &  72.8 & 64.5 & 67.7 & 58.6 & 71.1 & 72.0 & - \\
    LaSagnA-7B$^\dagger$~\cite{lasagna} & CLIP \text{\tiny ViT-L} + SAM & 13,052 & 76.8  & 78.7 &  73.8 & 64.4 & 70.6 & 60.1 & 70.6 & 71.9 & - \\
    Prompt-RIS$^*$~\cite{prompt_ris} & CLIP \text{\tiny ViT-B} + SAM  & - & 76.4  & 80.4 &  72.3 & 67.1 & 73.6 & 59.0 & 64.8 & 67.2 & 69.0 \\
    \hline
    Shared-RIS (Ours)  & BEiT-3 \tiny ViT-L  & 381 & \underline{77.0}  & \underline{78.8} &  \underline{74.7} & \underline{70.4} & \underline{75.1} & \underline{64.7} & \underline{71.9} & \underline{72.1} & \underline{69.7}  \\
    Shared-RIS (Ours)$^\ddagger$  & BEiT-3 \tiny ViT-L  & 381 & \textbf{82.8}  & \textbf{83.8} &  \textbf{80.6} & \textbf{75.9} & \textbf{78.5}  & \textbf{70.3} & \textbf{74.5} & \textbf{76.9} & \textbf{73.7}
    \\
    \hline
    \end{tabular}
    }
\label{tab:compare_large_scale_models}
    \vspace{-0.2cm}
\end{table*}

\subsection{Implementation Details}
\paragraph{Framework}
For a single encoder, we use a pre-trained BEiT-3 base model~\cite{beit3} with ViT-B~\cite{vit}.
This has the same configuration as ViT-B: $L=12$ layers, $p=16$ patch sizes, and $d=768$ channel dimension.
As in previous works~\cite{late_9_cgformer}, the input image size ($h \times w$) is set to ($480 \times 480$).
For a shared FPN, we use $K=4$ and $j=\{3,6,9,12\}$ intermediate stages of the encoder, with a reduced channel dimension of $d_w=512$.
For a shared decoder, the one shared transformer layer is used with $d_w=512$ channels.
We use a 0.35 value of a threshold $\tau$ to generate a binary mask from a probability map. 
For a loss function, two hyperparameters $\lambda_{\text{bce}}=2.0$ and $\lambda_{\text{dice}}=0.5$ are used.


\paragraph{Training Recipe}
Our model is optimized by AdamW optimizer~\cite{adamw} with $\beta_{1}=0.9$, $\beta_{2}=0.98$, and a weight decay of 5e-4. 
The learning rate is set to 1e-4 and decreases gradually according to a cosine annealing decay~\cite{cosine_warmup_decay}.
We use stochastic depth~\cite{stochastic_depth} with a probability of 0.1 and gradient clipping~\cite{gradnorm} with a 1.0 value.
The model is trained over 35 epochs with 14 batch sizes and 8 V100 GPUs.


\subsection{Dataset and Metric}
\paragraph{Dataset}
We evaluate our method using RefCOCO~\cite{refcoco}, RefCOCO+~\cite{refcoco}, and RefCOCOg~\cite{refcocog, refcocog_google} datasets.
RefCOCO, RefCOCO+, and RefCOCOg provide a total of 19,994, 19,992, and 26,711 images, respectively, and include 50,000, 49,856, and 54,822 instances annotated with multiple expressions.
Each dataset has unique characteristics in its textual form:
RefCOCO offers spatial descriptions with position terms such as ``\textit{left}'' and ``\textit{right}'', it is banned in RefCOCO+, and RefCOCOg provides lengthy and more complicated captions than others.

\paragraph{Metric}
We use the standard RIS metric: (1) oIoU (overall Intersection over Union), dividing the total intersection area by the total union area of all test samples, (2) mIoU (mean Intersection over Union), calculating the IoU for each individual sample and then averaging these values at final, and (3) prec@X, reporting the percentage of images with an IoU value above a threshold $\epsilon$, where $\epsilon \in \{ 0.5, 0.7, 0.9 \}.$


\subsection{Main Results}
\begin{figure*}[t]
     \centering
     \hspace{0.2cm}
     \begin{minipage}[b]{0.2\textwidth}
         \centering
         \includegraphics[width=\textwidth]{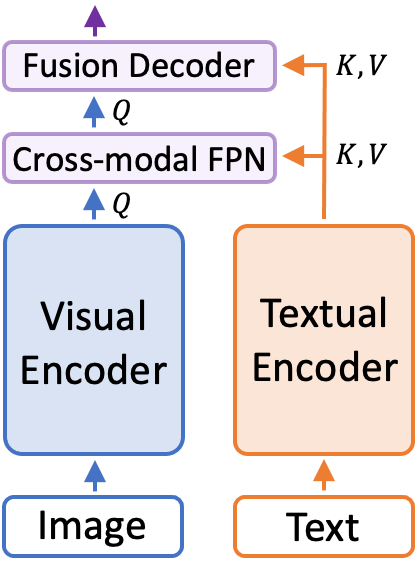}
         \caption*{(a) Late Fusion}
         \label{fig:late}
     \end{minipage}
     \hfill
     \begin{minipage}[b]{0.2\textwidth}
         \centering
         \includegraphics[width=\textwidth]{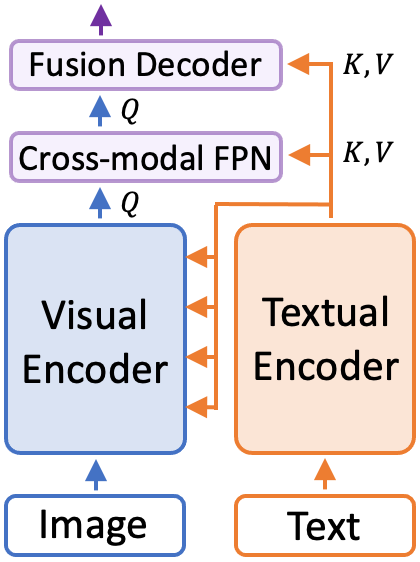}
         \caption*{(b) Early Fusion}
         \label{fig:early}
     \end{minipage}
     \hfill
     \begin{minipage}[b]{0.2\textwidth}
         \centering
         \includegraphics[width=\textwidth]{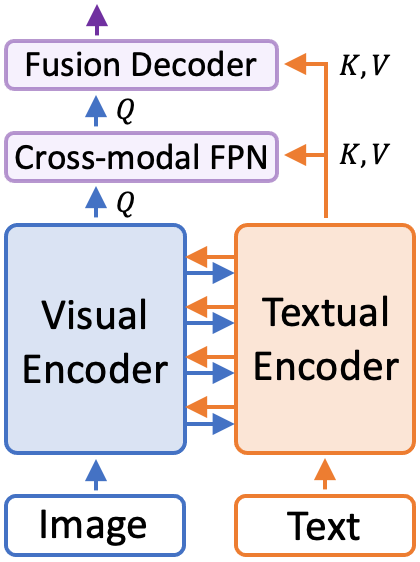}
         \caption*{(c) Joint Fusion}
         \label{fig:joint}
     \end{minipage}
     \hfill
     \begin{minipage}[b]{0.2\textwidth}
         \centering
         \includegraphics[width=\textwidth]{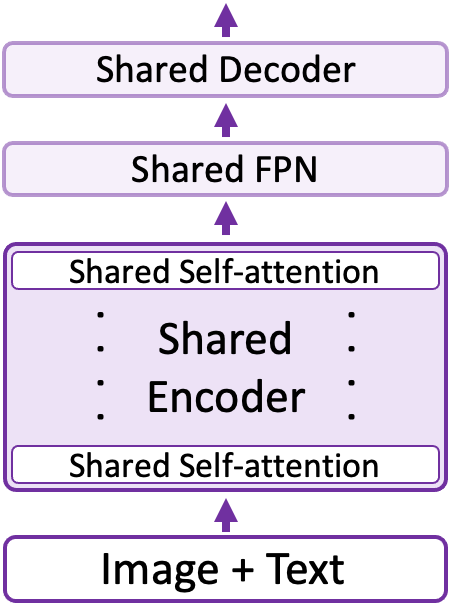}
         \caption*{(d) Seamless (Ours)}
         \label{fig:seamless}
     \end{minipage}
      \hspace{0.2cm}
        \caption{Illustration of fusion strategies in RIS: (a) late fusion, where final dense features are integrated at late phase, (b) early fusion, propagating text features at each stage of a visual encoder, (c) joint fusion, which allows joint interactions between two encoding stages. (a), (b), and (c) achieve partial interactions, limiting the achievable alignments from small scale of RIS data, whereas our seamless fusion in (d) continuously interacts two modalities within a shared model structure,
        leading to better alignments obtained after fine-tuning on RIS.
        }
        \label{fig:fusion_categories}
\end{figure*}

In Table~\ref{tab:main_results}, we compare our Shared-RIS with other referring segmentation methods on the oIoU metric and RIS benchmark datasets: RefCOCO, RefCOCO+, and RefCOCOg.
We also report the mIoU results in the supplementary materials.
The details of each fusion category (\eg, late, early, and joint fusions) are illustrated in Fig.~\ref{fig:fusion_categories}.
Our method achieves outstanding performance over the recent state-of-the-art RIS methods.
All compared methods are based on dual encoders.
The notable performance of our methods with a single encoder demonstrates that the mismatch in a desired multi-modal interaction level between backbone pre-training and a RIS task exists in current dual-encoders.
By aligning this, we obtain performance improvement with large margins.

In Table~\ref{tab:compare_large_scale_models}, we compare our methods with the RIS methods that leverage additional pre-trained models (\eg, LLM~\cite{llava} or SAM~\cite{sam}).
Our methods take disadvantages over these methods in terms of pre-trained data scales, as well as segmentation, and reasoning capabilities of SAM and LLM.
Moreover, LISA~\cite{lisa}, GSVA~\cite{gsva}, and LaSagnA~\cite{lasagna} use more segmentation data such as ADE-20k~\cite{ade20k} and COCO-Stuff~\cite{coco-stuff}, in addition to the combined RefCOCO variants.
Despite these disadvantages, our approach achieves the best performance with high efficiency.




Besides the performance results, the following three comparisons with each fusion category highlight the value of our Shared-RIS from different aspects.

\paragraph{Compared to Late Fusion Methods}
As illustrated in Fig.~\ref{fig:fusion_categories}a, late fusion approaches merge the final features from different encoders using their fusion decoder with unique characteristics.
For instance, (1) CRIS~\cite{late_3_cris} leverages the vision-language aligned features from CLIP to match visual pixel-level features with textual semantics.
Our approach significantly surpasses CRIS on all RefCOCO datasets, as shown in Table~\ref{tab:main_results}.
This demonstrates that our method achieves superior pixel-level alignments compared to those obtained by CRIS, utilizing CLIP.
This finding suggests that the insufficient interactions at CLIP's pre-training result in lower-level alignments than ours when fine-tuning on a RIS task.
(2) VLT~\cite{late_7_vlt} achieves the second best performance on the test (U) and val (G) split of RefCOCOg which require multi-modal reasoning ability to a RIS model due to its lengthy and complex textual forms.
They aim to improve joint reasoning ability via multiple query set generations, leading them to second-best results on RefCOCOg.
By outperforming VLT with large margins on RefCOCOg, we validate that our shared-fusion approach is more capable of handling complicated correlations in visual contexts and textual semantics than VLT.
(3) Our method also outperforms CGFormer~\cite{late_9_cgformer} and ReLA~\cite{late_1_ReLA}, which are specifically designed to capture object-level relationships.
This highlights the ability of our Shared-RIS to manage intricate object relations better.

\paragraph{Compared to Early Fusion Methods}
We also demonstrate the effectiveness of our seamless fusion approach compared with early fusion methods~\cite{early_1_lavt,early_3_dmmi}, where they directly integrate linguistic semantics into visual encoding phrases with cross-attention modules to align multi-modal features densely, as depicted in Fig.~\ref{fig:fusion_categories}b.
We outperform all early fusion methods, as in Table~\ref{tab:main_results}, showing the superiority of our seamless interactions across a framework in producing finer aligned multi-modal features than early fusions.

\paragraph{Compared to Joint Fusion Methods}
The methods with joint fusions~\cite{joint_3_crossvlt, joint_2_coupalign, joint_4_risclip} aim at more fine-grained alignments than early fusion methods by allowing cross-modal interaction at the intermediate layers of two encoders, jointly.
Compared to them, our seamless fusion achieves noticeable performance improvements, as in Table~\ref{tab:main_results}.
These results suggest that our methods, which start deeper interactions with shared fusions from the initial inputs, are simple and more effective in aligning multi-modal features densely, than starting interactions from the first stage of two distinct encoders like joint fusion methods.

\begin{figure}[t]
    \centering
    \begin{minipage}[b]{1\linewidth}
        \centering
        \includegraphics[width=\linewidth]{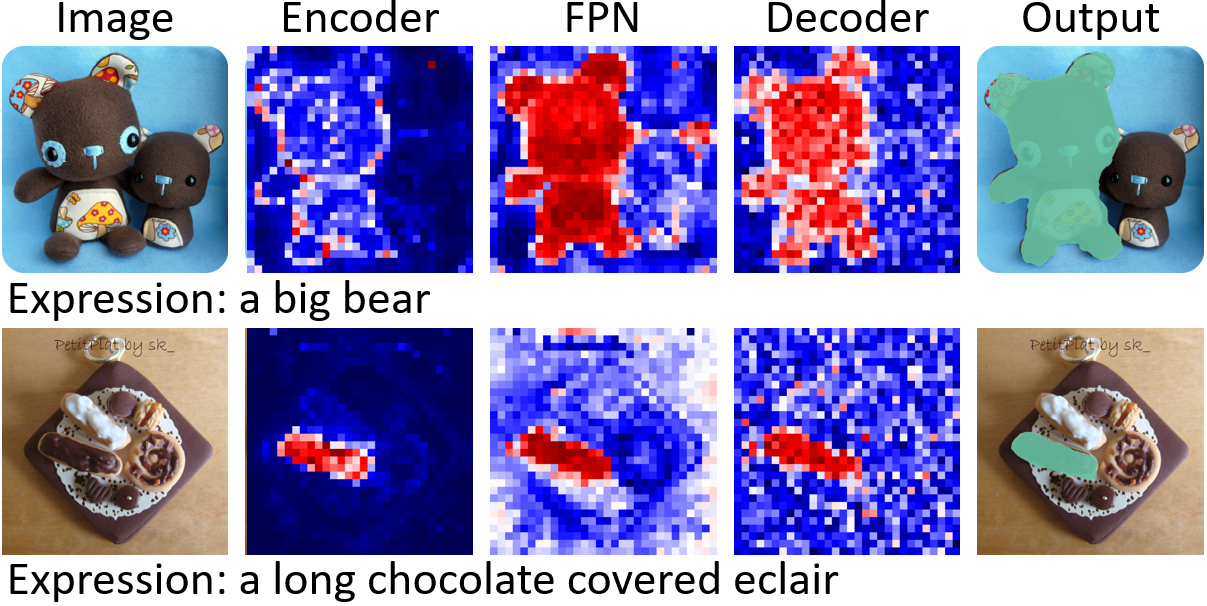}
        \caption{Visualization of attention maps on each component of our Shared-RIS.}
        \vspace{0.2cm}
        \label{fig:attn_component}
    \end{minipage}
    \begin{minipage}[b]{1\linewidth}
        \centering
        \includegraphics[width=\linewidth]{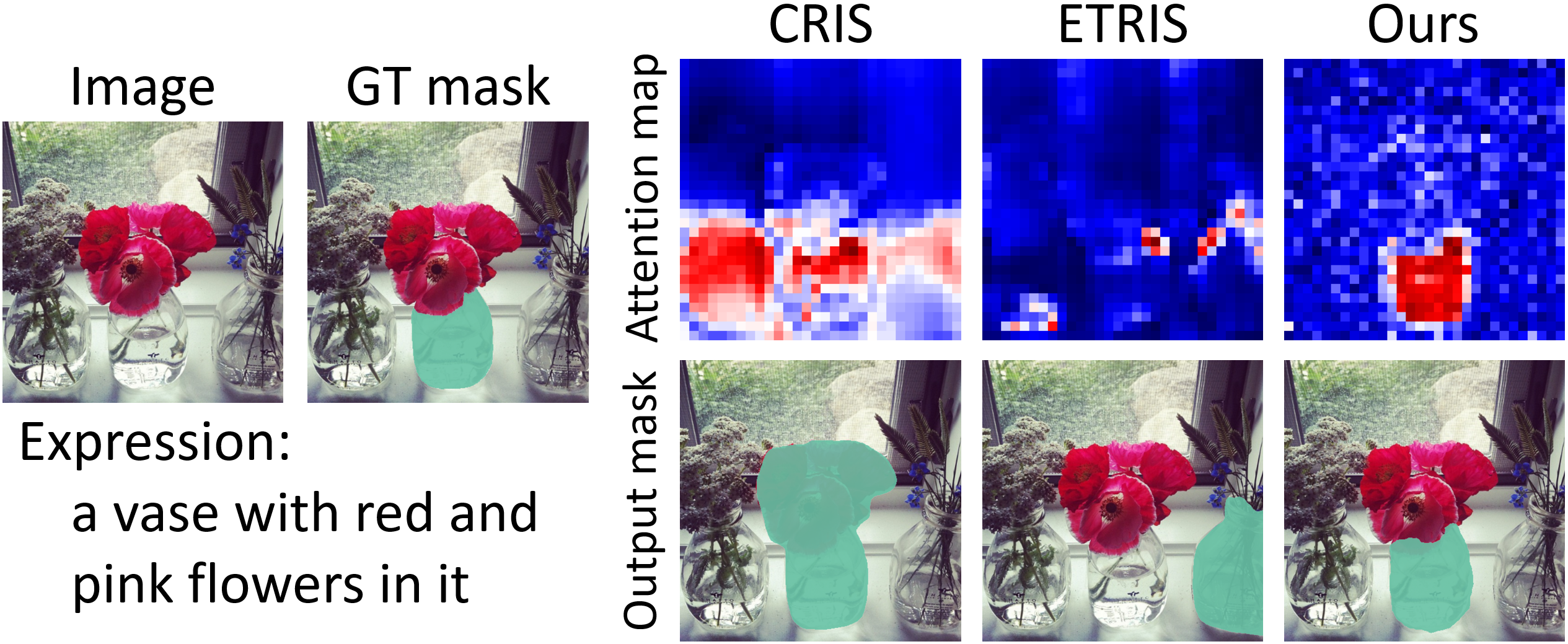}
        \caption{Visualization of attention maps compared to other RIS methods.}
        \label{fig:attn_compare}
        \vspace{0.2cm}
    \end{minipage}
    \begin{minipage}[b]{1\linewidth}
        \centering
        \includegraphics[width=\linewidth]{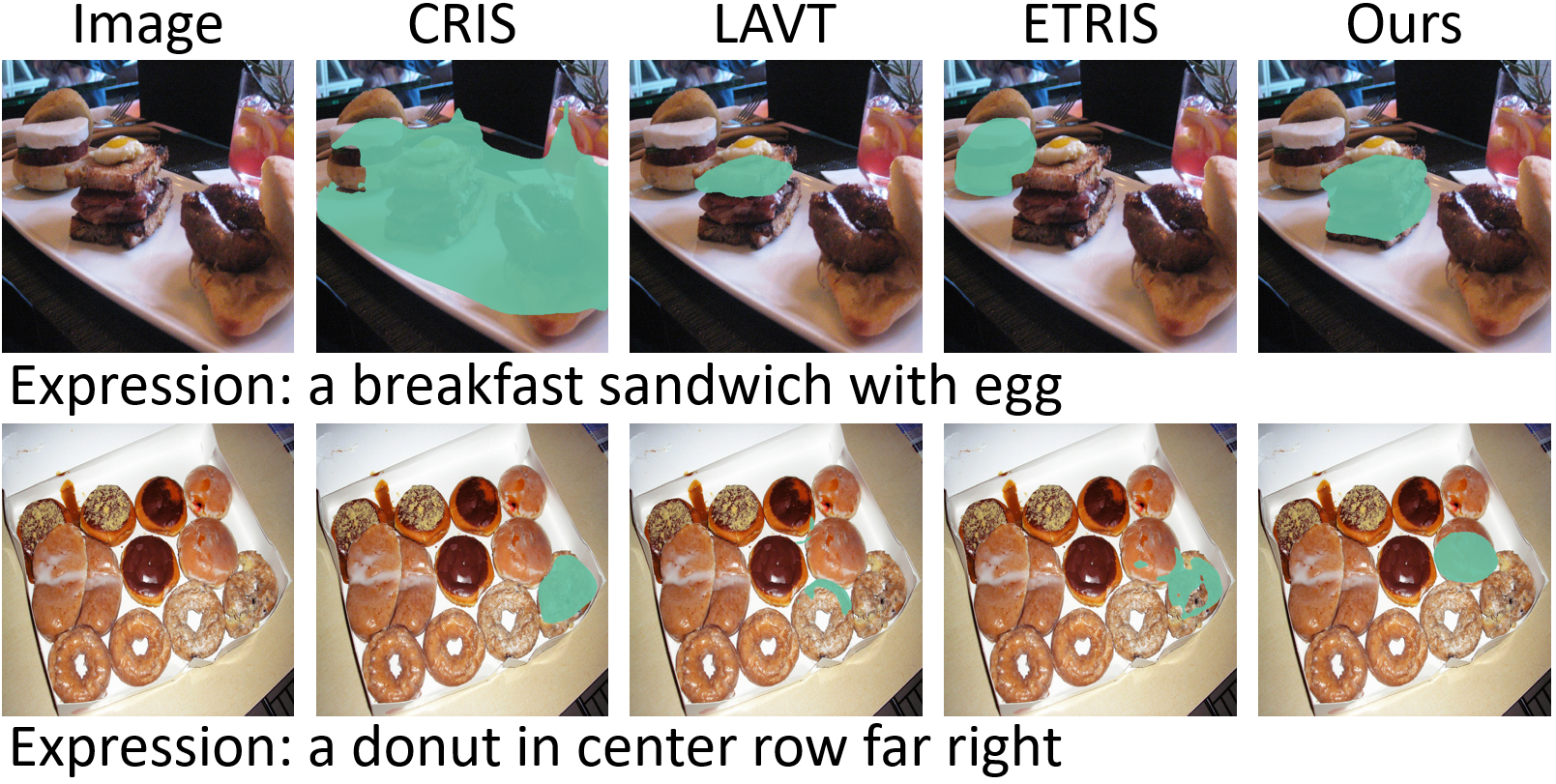}
        \caption{Qualitative analysis of Shared-RIS compared to other RIS methods.}
        \label{fig:qual}
        \vspace{-0.6cm}
    \end{minipage}
\end{figure}

\paragraph{Qualitative Analysis}
Fig.~\ref{fig:qual} shows the segmentation mask generated by our methods compared with other fusion strategies including late fusion~\cite{late_3_cris}, early fusion~\cite{early_1_lavt}, and joint fusion~\cite{joint_1_etris}.
Specifically, the expression in (a): our methods identify ``\textit{egg}'' word that offers a critical cue distinguishing a target sandwich from other sandwiches and (b): demonstrating the ability of our model to handle spatial terms such as ``\textit{center row}''. 
However, the compared methods fail to distinguish the referred target from visually similar objects.
More qualitative results with diverse situations are presented in our supplementary. 

\paragraph{Attention Map Analysis}
Fig.~\ref{fig:attn_component} presents the attention maps for each component of our framework, including a shared encoder, FPN, and decoder.
These attention maps consistently highlight the target regions at all components, making well-aligned visual and textual features. 
Fig.~\ref{fig:attn_compare} compares the attention maps generated by ours with those by other RIS methods~\cite{late_3_cris, joint_1_etris}.
The attention maps in all compared methods fail to accurately focus on a target object, whereas ours have high attention values along with target regions.
These visualizations of attention maps confirm that shared self-attention is effective in aligning visual regions with textual semantics for RIS.
The detailed process of these visualizations is detailed in our supplementary.

\subsection{Ablation Study}
We conduct ablation studies on the validation set of RefCOCO+ to prove the effectiveness of our framework.
We also compare ours with other RIS methods in terms of parameters and FLOPs (measured by fvcore library\footnote{https://github.com/facebookresearch/fvcore}) to show our efficiency.

\paragraph{Impact of Proposed Components}
In Table~\ref{tab:ablation_components}, we analyze the contribution of each proposed component within our framework.
We observe that na\"ively using a single encoder achieves competitive performance, indicating that it is a suitable backbone for a RIS task and supporting our motivation.
Integrating the proposed FPN or decoder into a single encoder significantly improves performance across all metrics.
Notably, combining both shared FPN and shared decoder leads to further improvements, implying their synergistic effect.

    

\begin{table}[t]
    \centering
    \small
    \caption{Ablation study with different components. No checkmark $\checkmark$ on the \textit{Shared Decoder} column indicates directly predicting segmentation mask without a shared transformer layer in a shared decoder.}
    \scalebox{1.0}{
    \begin{tabular}{c@{\hspace{7pt}}c@{\hspace{7pt}}|l@{\hspace{9pt}}l@{\hspace{9pt}}l@{\hspace{9pt}}l@{\hspace{9pt}}l}
    \hline
    \multirow{2}{*}{\begin{tabular}[c]{@{}c@{}}Shared\\ FPN\end{tabular}} & \multirow{2}{*}{\begin{tabular}[c]{@{}c@{}}Shared\\ Decoder\end{tabular}} & \multirow{2}{*}{P@0.5} & \multirow{2}{*}{P@0.7} & \multirow{2}{*}{P@0.9} & \multirow{2}{*}{oIoU} & \multirow{2}{*}{mIoU} \\ 
    & & & & \\
    \hline
        &     & 79.36 & 64.70 & 10.16 & 65.12 & 66.40 \\
    \checkmark & & 81.87 & 70.25 & 13.59 & 67.31 & 69.10  \\
    & \checkmark & 81.39  & 69.53  & 13.42 & 66.25 & 68.38  \\
    \checkmark & \checkmark & \textbf{82.77}  & \textbf{72.12}  & \textbf{16.87} & \textbf{68.42} & \textbf{70.34} \\ \hline
    
    \end{tabular}
    }    \label{tab:ablation_components}
    \vspace{-0.1cm}
\end{table}

\begin{table}[t]
    \centering
    \small
    \caption{Comparisons with other FPN methods within our framework, where our shared decoder is applied.}
    \scalebox{1.00}{
    \begin{tabular}{l@{\hspace{7pt}}c@{\hspace{7pt}}c@{\hspace{7pt}}|c@{\hspace{9pt}}c}
    \hline
    FPN Methods & Params & GFLOPs & oIoU & mIoU \\
    \hline
    No FPN & - & - & 66.25 & 68.38 \\
    FPN in CRIS~\cite{late_3_cris} & 23M & 19 & 66.91 & 68.37 \\
    PWAM in LAVT~\cite{early_1_lavt} & 14M & 13 & 67.49 & 69.77 \\
    \cdashline{1-5}[1pt/1pt]
    Shared FPN (Ours) & \textbf{8M} & \textbf{9} & \textbf{68.42} & \textbf{70.34} \\
    \hline
    \end{tabular}
    }
    \label{tab:fpn}
\end{table}

\begin{table}[t]
    \centering
    \small
    \caption{Comparisons with other fusion decoder methods within our Shared-RIS, where our shared FPN is employed. $\dagger$ denotes directly generating segmentation mask without shared fusion in our decoder.}
    \scalebox{1.00}{
    \begin{tabular}{l@{\hspace{7pt}}c@{\hspace{7pt}}c@{\hspace{7pt}}|c@{\hspace{9pt}}c}
    \hline
    Decoder Methods & Params & GFLOPs & oIoU & mIoU \\
    \hline
    No Fusion$^\dagger$ & \textcolor{gray}{5M} & \textcolor{gray}{46} & 67.31 & 67.71 \\
    CGFormer~\cite{late_9_cgformer} & 44M & 857 & 67.40 & 69.07\\
    DMMI~\cite{early_3_dmmi} & 94M & 276 & 67.70 & 69.10 \\
    \cdashline{1-5}[1pt/1pt]
    Shared Decoder (Ours) &  \textbf{9M} & \textbf{50} & \textbf{68.42}  & \textbf{70.34}  \\
    \hline
    \end{tabular}
    }
    \label{tab:decoder}
\end{table}

\begin{table}[t]
    \centering
    \small
    \caption{
    Comparisons on efficiency and performance with the recent SoTA methods using a single-backbone or not. $\dagger$ indicates that public codes are not available.
    }
    \scalebox{0.92}{
    \begin{tabular}{l@{\hspace{5pt}}|l@{\hspace{5pt}}c@{\hspace{5pt}}c@{\hspace{5pt}}|c@{\hspace{5pt}}c}
    \hline
    Methods & Backbones   & Params & GFLOPs & oIoU & mIoU \\ \hline
    \multirow{2}{*}{RISCLIP$^\dagger$~\cite{joint_4_risclip}} & \multirow{2}{*}{CLIP + CLIP} & \multirow{2}{*}{375M} & \multirow{2}{*}{1380} & \multirow{2}{*}{65.53} & \multirow{2}{*}{69.16} \\
           & & & & & \\
           \cdashline{1-6}[1pt/1pt]
                   \cdashline{1-6}[1pt/1pt]

    \multirow{2}{*}{CGFormer~\cite{late_9_cgformer}} & Swin + BERT & 252M & 949 & 64.54 & 68.56  \\ 
     & BEiT-3 & 269M & 955  & 67.44 & 69.10 \\ 
    
   \cdashline{1-6}[1pt/1pt]
    \multirow{2}{*}{DMMI~\cite{early_3_dmmi}} & Swin + BERT & 341M & 392 & 63.98 & 67.51 \\ 
     & BEiT-3 & 356M & 397 & 67.64 & 68.97 \\ 
        \hline
    Shared-RIS (Ours) & BEiT-3 & \textbf{239M} &\textbf{155} & \textbf{68.42} & \textbf{70.34} \\  
    \hline
    \end{tabular}
    }
    \label{tab:others_on_single}
    \vspace{-0.1cm}
\end{table}

\paragraph{Comparison with other FPNs}
Table~\ref{tab:fpn} validates our shared FPN by comparing it with other FPN methods (\eg a proposed FPN in CRIS~\cite{late_3_cris} and PWAM in LAVT~\cite{early_1_lavt}) in two aspects: efficiency (parameters and FLOPs) and performance (oIoU and mIoU).
These comparisons are made by adapting different FPN methods into our framework.
We notice that the proposed FPN obtains higher performance than all compared FPNs with lower parameters and FLOPs, implying that a shared fusion layer alone effectively learns detailed features for RIS without a sophisticated design that increases complexity.

\paragraph{Comparison with other Fusion Decoders}
In Table~\ref{tab:decoder}, we show that our shared decoder outperforms other fusion decoders~\cite{early_3_dmmi, late_9_cgformer} in terms of both performance and computational cost, by integrating them into our framework.
The compared decoders demand high computational costs (in terms of both parameters and FLOPs) because they (1) consist of multiple transformer decoder layers, (2) utilize high-resolution features from the visual encoder in their decoders, and (3) have complicated designs to improve performance over prior methods.
In contrast, our decoder achieves superior performance even with a single shared transformer layer, resulting in a more lightweight and effective method, as evidenced by our fewer parameters and lower FLOPs.
These results suggest that a shared self-attention mechanism, which encodes two modalities together, is valuable for a RIS task.


\paragraph{Others on Single-encoder}
In this ablation study, by adapting a single backbone to existing SoTA methods, we highlight that (1) a single-encoder is the optimal backbone for RIS over dual encoders, and (2) our design, which maximizes the capabilities of shared self-attention, is effective.
Table~\ref{tab:others_on_single} shows that a single backbone significantly improves the performance of the compared methods~\cite{early_3_dmmi, late_9_cgformer}.
Notably, our framework outperforms all compared methods, validating that shared self-attention is a well-suited fusion method for RIS to align two modalities.

\paragraph{Efficiency Comparison}
In Table~\ref{tab:others_on_single}, our Shared-RIS shows the best efficiency in both parameters and FLOPs compared to recent SoTA methods~\cite{joint_4_risclip, early_3_dmmi, late_9_cgformer} with large margins.
When adapting a single-encoder to other methods, their complexities slightly increase.
Our FPN and decoder highly contribute to the lightweight nature of our model with the best performance.


\paragraph{Ablation on Shared FPN}
We conducted an ablation study within our Shared FPN, evaluating various intermediate stages for multi-level feature maps in Table~\ref{tab:supple_ablation_fpn}(a) and different fusion modules in Table~\ref{tab:supple_ablation_decoder}(b). 
Compared to the baseline without FPN ("No FPN"), 
\begin{wrapfigure}{r}{0.19\textwidth}
\vspace{-0.3cm}
    \centering
    \includegraphics[width=0.19\textwidth]{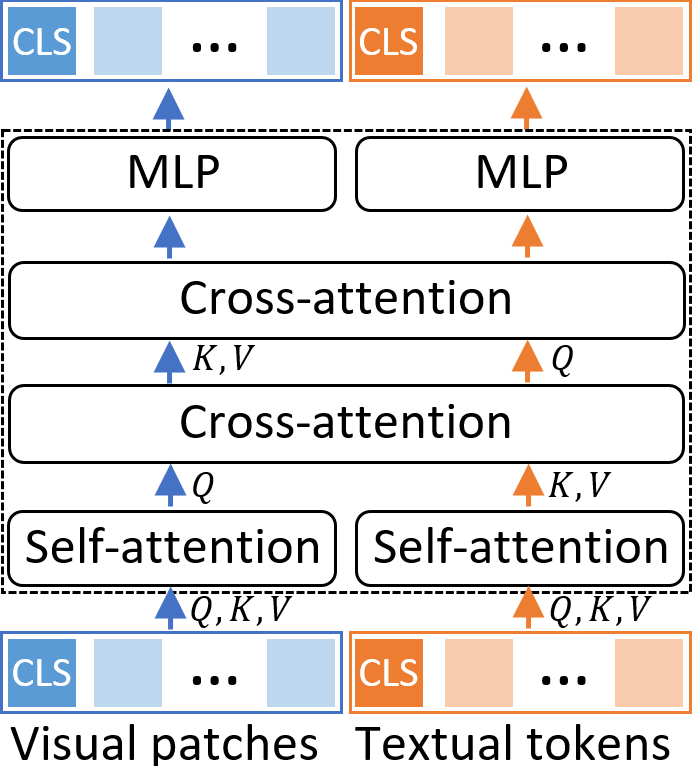}
    \vspace{-0.5cm}
    \caption{Bi-direction.}
    \label{sup_fig:bi-direction}
\vspace{-0.4cm}
\end{wrapfigure}
all variations in stages significantly improved performance, with $K=4$ intermediate stages at ${3,6,8,12}$ yielding the best results.
The self-attention in our FPN outperforms cross-attention and bi-directional methods (proposed in Grounding DINO~\cite{grounding_dino} and depicted in Fig.~\ref{sup_fig:bi-direction}), with fewer parameters.
This highlights the effectiveness of our self-attention design for RIS.


\begin{table}[t]
    \centering
    \small
    \caption{Ablation on Shared FPN, where a shared decoder is employed. $\dagger$ denotes the module proposed in Grounding DINO and is illustrated in Fig.~\ref{sup_fig:bi-direction}.}
    \scalebox{1.0}{
    \begin{tabular}{l|c@{\hspace{5pt}}c@{\hspace{5pt}}|c@{\hspace{5pt}}c@{\hspace{5pt}}}
    \hline
    Ablations   & Params & GFLOPs & oIoU & mIoU \\
    \hline
    \multicolumn{5}{@{\hspace{12pt}}l}{(a) \textit{Intermediate Stages} } \\ \cdashline{1-5}[1pt/1pt]
    No FPN & - & - & 66.25 & 68.38 \\
    $ \{ $6, 12$\}$       & 4M & 5 & 67.74  &  69.89 \\
    $ \{ $4, 8, 12$\}$       & 6M & 7  & 68.13  & 70.24   \\ 
    \textbf{$ \{ $3, 6, 9, 12$\}$ }      & 8M & 9 & \textbf{68.42} & \textbf{70.34} \\
    \hline
    \hline
    \multicolumn{5}{@{\hspace{12pt}}l}{(b) \textit{Fusion Modules}} \\ \cdashline{1-5}[1pt/1pt]
    No FPN & - & - & 66.25 & 68.38 \\
    Cross-attention & 8M & 6  & 66.98  & 69.20 \\
    Bi-direction$^{\dagger}$ & 37M & 22 & 68.19 & 70.02 \\
    \textbf{Shared self-attention}  & 8M & 9  & \textbf{68.42} & \textbf{70.34} \\
    \hline
    \end{tabular}
    }
    \label{tab:supple_ablation_fpn}
\end{table}
\begin{table}[t]
    \centering
    \small
    \caption{Ablation on Shared Decoder, where a shared FPN is employed. $\dagger$ denotes the module proposed in Grounding DINO and is illustrated in Fig.~\ref{sup_fig:bi-direction}}
    \scalebox{1.0}{
    \begin{tabular}{l|c@{\hspace{5pt}}c@{\hspace{5pt}}|c@{\hspace{5pt}}c@{\hspace{5pt}}}
    \hline
    Ablations   & Params & GFLOPs & oIoU & mIoU \\

    \hline
    \multicolumn{3}{@{\hspace{12pt}}l}{ (a) \textit{Shared Transformer Layers}} \\ \cdashline{1-5}[1pt/1pt]
    No Fusion & \textcolor{gray}{5M} & \textcolor{gray}{46} & 67.31 & 67.71 \\
    \textbf{1} (Ours)     & \textbf{9M} & \textbf{50} & 68.42 & 70.34 \\
    2       & 12M & 54 & \textbf{68.57}  & \textbf{70.59}   \\ 
    3       & 15M & 58 & 68.27  & 70.29   \\
    \hline
    \hline
    \multicolumn{3}{@{\hspace{12pt}}l}{(b) \textit{Fusion Modules}} \\ \cdashline{1-5}[1pt/1pt]
    No Fusion & \textcolor{gray}{5M} & \textcolor{gray}{46} & 67.31 & 67.71 \\
    Cross-attention & 9M & 48 & 67.29 & 69.30     \\
    Bi-direction$^{\dagger}$ & 14M & 51 & 68.30 & 70.09 \\
    \textbf{Shared self-attention} & 9M & 50 & \textbf{68.42}  & \textbf{70.34} \\
    \hline
    \end{tabular}
    }
    \vspace{-0.2cm}
    \label{tab:supple_ablation_decoder}
\end{table}

\begin{figure}
    \vspace{-0.3cm}
    \centering
    \includegraphics[width=0.95\linewidth]{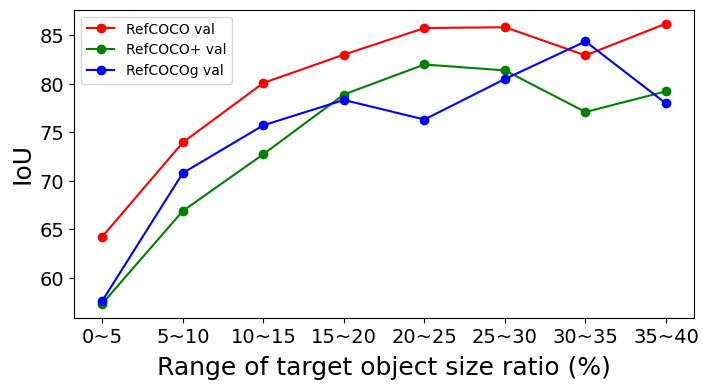}
    \caption{Anaysis on IoU performance of our methods depending on target object size ratio (\%), which is calculated by dividing the target object area by the image area. Our methods have limitations on small objects.}
    \label{fig:object_size}
\end{figure}

\paragraph{Ablation on Shared Decoder}
\label{appendix:ablation_decoder}
We also conduct an ablation study on our shared decoder with the effects of varying the number of shared transformer layers in Table~\ref{tab:supple_ablation_decoder}(a), and the types of fusion modules in Table~\ref{tab:supple_ablation_decoder}(b).
For part (a), we observe that using two shared transformer layers yields the highest performance, while increasing the complexity more than one layer.
We therefore choose one shared layer as our final decoder to balance performance and efficiency.
In part (b), we discover that using shared self-attention for the interaction of two modalities in a decoder surpasses the effectiveness of cross-attention and bi-direction modules proposed in Grounding DINO~\cite{grounding_dino} and shown in Fig.~\ref{sup_fig:bi-direction}.
These results demonstrate the superior capability of shared self-attention in learning multi-modal representations for the RIS task.


\section{Limitation and Discussion}
While our approach excels in capturing fine-grained details for RIS, it is less effective when dealing with small objects.
Fig.~\ref{fig:object_size} illustrates the IoU performance of our methods depending on the target object size ratio (\%), which is calculated by dividing the target object area by the image area.
These results show that IoU is relatively lower for small objects compared to larger ones.
This challenge arises because our methods do not utilize hierarchical features (\ie, multi-scale features) inherent in the plain ViT architecture of BEiT-3.
We plan to explore and analyze this matter further in the future.


Our method also inherits the negative social impacts of existing RIS models, such as biases in web-sourced pre-training data and manually annotated RIS datasets, potentially leading to unfair and discriminatory results.

\section{Conclusion}
\label{sec:conclusion}

In this paper, we present a simple baseline with single-encoder that fully exploits the capabilities of shared self-attention, featuring a lightweight FPN and mask decoder.
We also provide three insightful findings:
(1) Aligning multi-modal interaction levels between backbone pre-training and a RIS task is valuable,
(2) deeper and seamless interactions across all processes are effective for pixel-word alignments,
and (3) shared self-attention is a very promising fusion method for RIS.
These findings drive our framework to achieve outstanding performance on the RIS benchmark with high efficiency.

\bibliographystyle{IEEEtran}
\bibliography{custom_arxiv}

\clearpage

\appendix
\label{sec:appendix}

\begin{table*}[t!]
    \centering
    \small
    \caption{mIoU comparison with other RIS methods on RefCOCO, RefCOCO+, and RefCOCOg.
    $\dagger$ denotes the model trained on the combined RefCOCO datasets.
    }
    \scalebox{1.00}{
    \begin{tabular}{l|l|l|ccc|ccc|ccc}
    \hline
    \multirow{2}{*}{Methods}  & \multicolumn{2}{c|}{Encoders}& \multicolumn{3}{c|}{RefCOCO}& \multicolumn{3}{c|}{RefCOCO+} & \multicolumn{3}{c}{RefCOCOg} \\ \cline{2-12}
         & Visual & Textual & val & testA & testB & val & testA & testB & val(U) & test(U) & val(G) \\ \hline
    CRIS~\cite{late_3_cris} & CLIP \tiny R101 & CLIP & 70.34 & 73.41 & 65.72 & 62.13 & 67.68 & 53.42 & 60.30 & 60.29 & - \\
     ETRIS~\cite{joint_1_etris} & CLIP \tiny R101 & CLIP & 71.06 & 74.11 & 66.66 & 62.23 & 68.51 & 52.79 & 60.28 & 60.42 & 57.86 \\
     LAVT~\cite{early_1_lavt} & Swin-B & BERT & 74.46 &	76.89&	70.94	&65.81&	70.97	&59.23	&63.34	&63.62&	63.66 \\
     DMMI~\cite{early_3_dmmi} & Swin-B & BERT & 75.26 & 76.96 & 72.05 & 67.51 & 72.10 & 60.38 & 66.48 & 67.07 & - \\
     CrossVLT~\cite{joint_3_crossvlt} & Swin-B & BERT & 75.48 & 77.54 & 72.69 & 67.27 & 72.00 & 60.09 & 66.21 & 62.09 & - \\
     RISCLIP~\cite{joint_4_risclip} & CLIP \tiny ViT-B & CLIP & \textbf{75.68} & \textbf{78.01} & 72.46 & 69.16 & 73.53 & 60.68 & 67.61 & 67.95 & - \\
     \hline
    Shared-RIS (Ours)  & \multicolumn{2}{c|}{BEiT-3 \tiny ViT-B} & 75.50 & 76.66 &\textbf{73.03} & \textbf{70.34} & \textbf{73.75} & \textbf{65.07} & \textbf{68.50} & \textbf{69.17} & \textbf{66.65} \\
        Shared-RIS (Ours)  & \multicolumn{2}{c|}{BEiT-3 \tiny ViT-L} & \textcolor{gray}{78.09}  & \textcolor{gray}{79.30} &  \textcolor{gray}{76.33} & \textcolor{gray}{73.70} & \textcolor{gray}{76.65} & \textcolor{gray}{68.88} & \textcolor{gray}{71.88} & \textcolor{gray}{72.16}  & \textcolor{gray}{71.14} \\
    Shared-RIS (Ours)$^\dagger$  & \multicolumn{2}{c|}{BEiT-3 \tiny ViT-L} & \textcolor{gray}{82.73}  & \textcolor{gray}{83.44} &  \textcolor{gray}{81.6} & \textcolor{gray}{78.13} & \textcolor{gray}{80.07} & \textcolor{gray}{74.12} & \textcolor{gray}{75.34} & \textcolor{gray}{77.34}  & \textcolor{gray}{76.58} \\
    \hline
    \end{tabular}
    }
    \label{suptab:miou_results}
    \vspace{-0.2cm}
\end{table*}

\section*{mIoU Results.}
We report mIoU results of our methods compared to other RIS methods in Table~\ref{suptab:miou_results}.
Our counterpart, \ie, RISCLIP~\cite{joint_4_risclip}, shows the higher performance on the val and testA split of RefCOCO dataset, but they demand remarkable computational costs in both parameters and FLOPs than our Shared-RIS, as reported in Table VI of the main manuscript.
By outperforming RISCLIP with lower computational overhead on all other dataset splits, we show the efficiency of our approach. 
\label{appendix:miou}

\section*{Ablation on Segmentation Frameworks}
\label{appendix:ablation_framework}
To highlight the value of our Shared-RIS, we compare the performance of an instance segmentation framework (ViT-DET~\cite{vit_det}) reported in BEiT-3 on RefCOCO+ with our framework in Table~\ref{tab:supple_ablation_framework}.
For this comparison, we modify the number of classes in a prediction head of ViT-DET into a two-way classification (\eg, background and foreground), instead of COCO 80 classes, to align with the RIS task.
Our framework outperforms the ViT-DET with BEiT-3 framework, demonstrating that (1) the success of our approach is not solely attributed to the capabilities of BEiT-3, and (2) our simple baseline is a well-designed architecture for RIS.

\begin{table}[h]
    \centering
    \small
    \caption{Ablation on Framework. $\dagger$ denotes the na\"ive extension of the instance segmentation framework used in BEiT-3 into a RIS framework.}
    \scalebox{1.1}{
    \begin{tabular}{l|l|c@{\hspace{7pt}}c@{\hspace{7pt}}}
    \hline
    Backbone & Framework    & oIoU & mIoU \\
    \hline
      \multirow{2}{*}{BEiT-3} & ViT-DET$^{\dagger}$~\cite{vit_det}  & {67.23} & {68.82} \\
      \cdashline{2-4}[1pt/1pt]
      & Shared-RIS (Ours)    & {\textbf{68.42}} & {\textbf{70.34}} \\
    \hline
    \end{tabular}
    }
    \label{tab:supple_ablation_framework}
\end{table}

\section*{More Visualizations}
\label{appendix:more_qual}
\begin{figure*}[t!]
    \centering
    \includegraphics[width=0.95\linewidth]{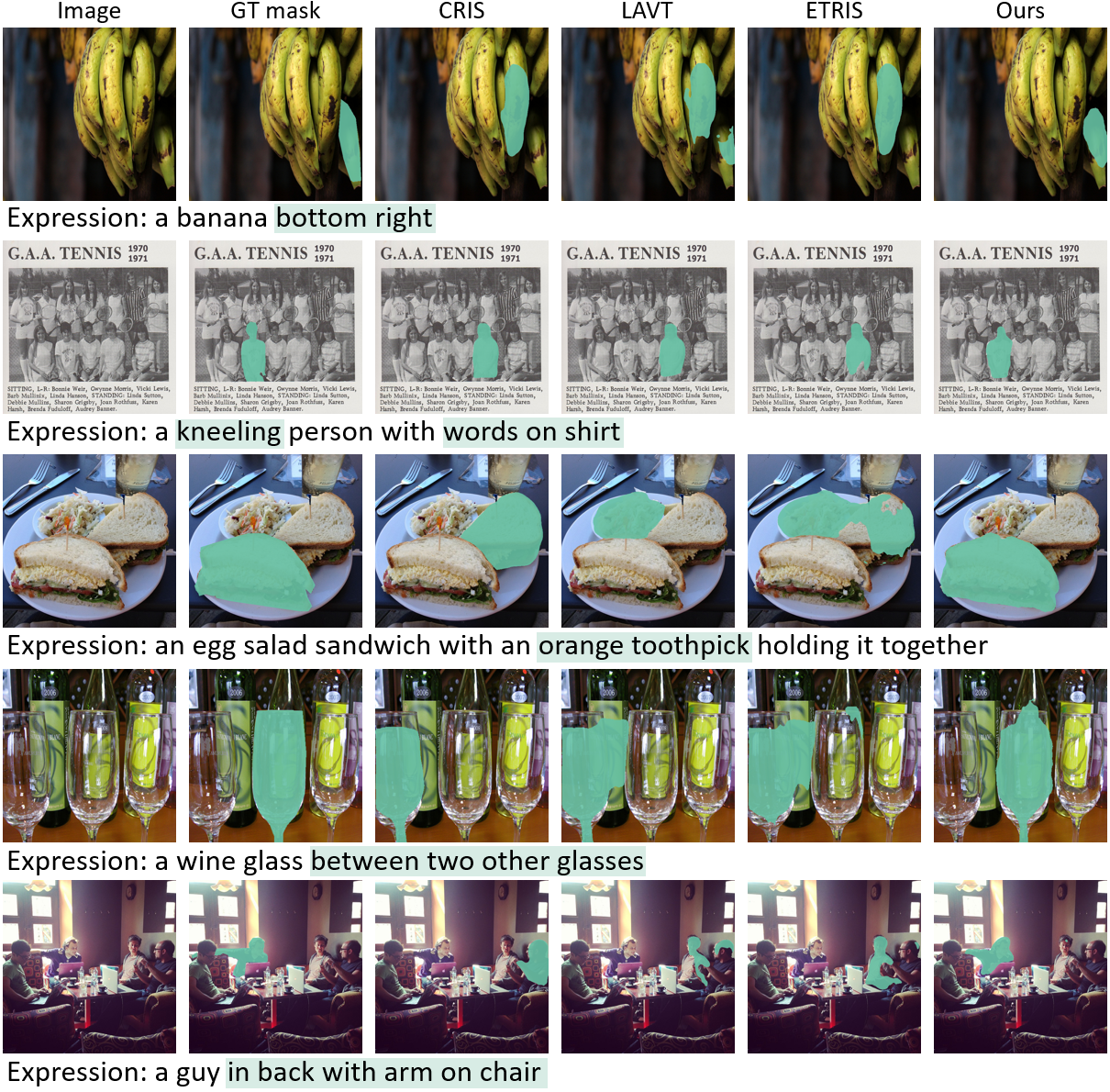}
    \caption{More qualitative analysis of Shared-RIS compared to other fusion strategies. We emphasize keywords with a colored background which provide the critical cues distinguishing a target from other similar ones.}
    \label{fig:supple_fig_qual}
\end{figure*}

\paragraph{Qualitative Results}
In Fig.~\ref{fig:supple_fig_qual}, we provide qualitative comparisons with other fusion strategies~\cite{late_3_cris, early_1_lavt, joint_1_etris}, validating the superiority of our Shared-RIS in more diverse situations.
We highlight the keywords with a colored background to emphasize the words that are crucial for distinguishing a target from other similar objects.

\begin{figure}[t]
    \centering
    \includegraphics[width=1\linewidth]{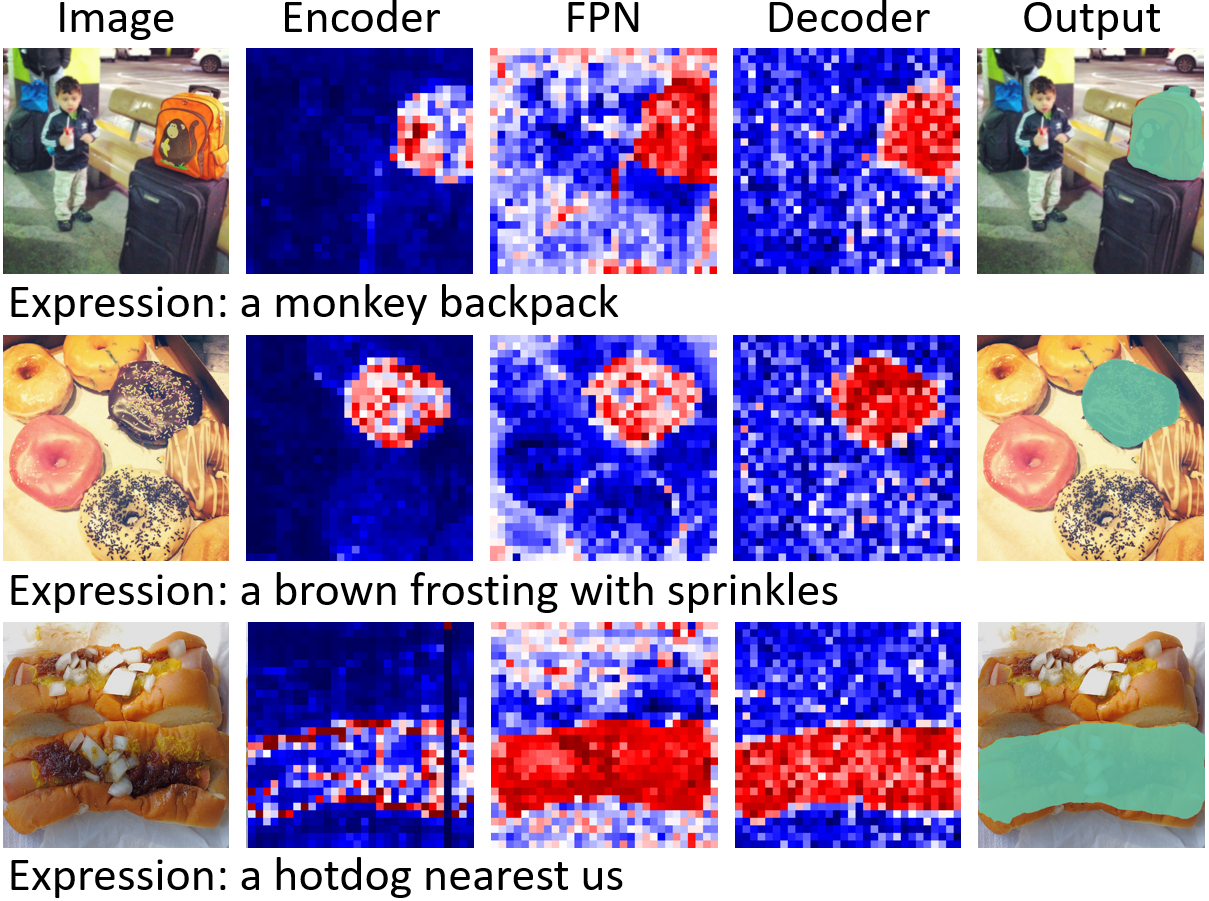}
    \caption{More visualization of attention maps for each component of our Shared-RIS.}
    \vspace{-0.2cm}
    \label{fig:supple_fig_attention_map}
\end{figure}

\begin{figure}[t]
    \centering
    \includegraphics[width=1\linewidth]{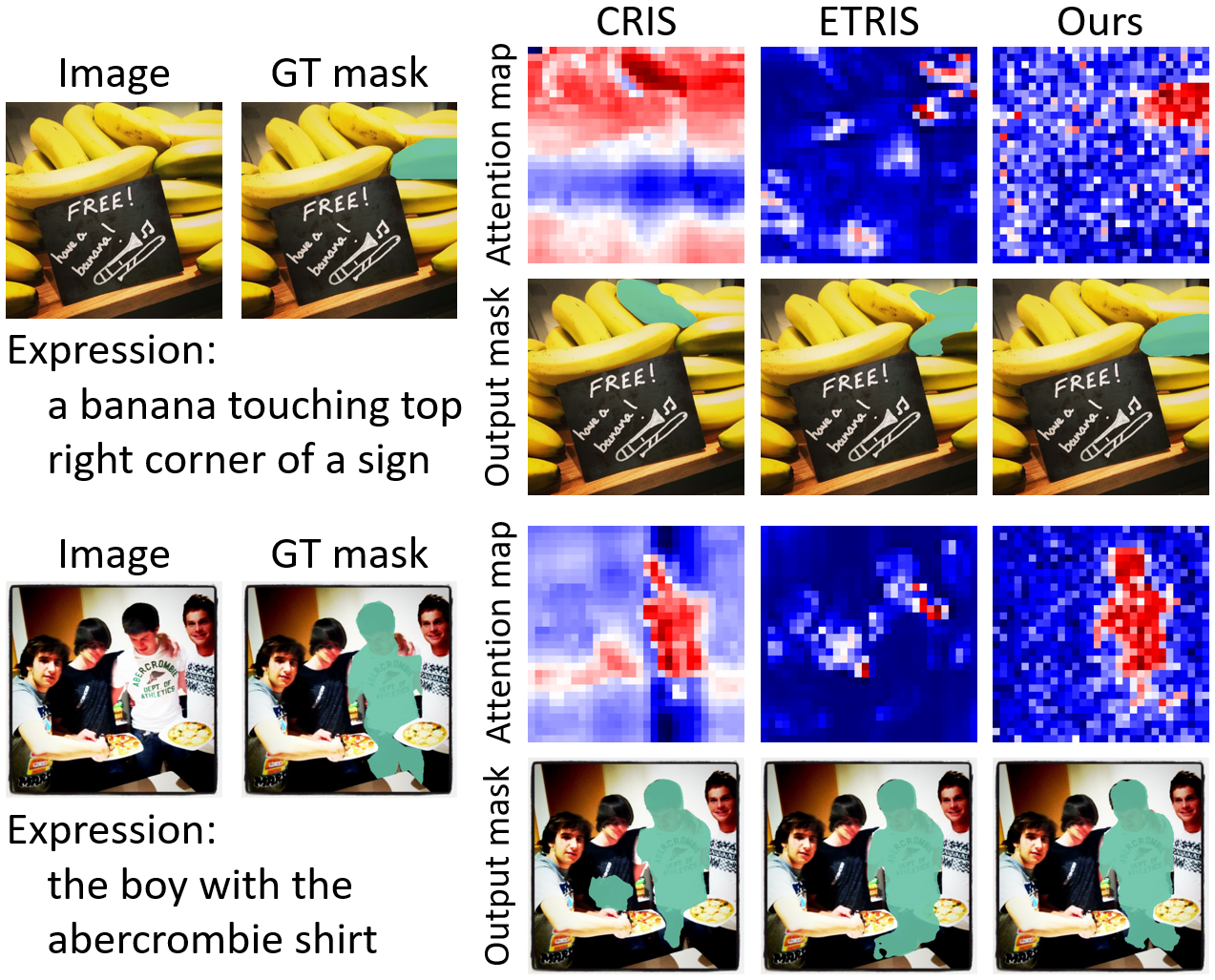}
    \caption{More visualization of attention maps compared to other RIS methods.}
    \label{fig:supple_fig_attention_map_sotas}
\end{figure}

\paragraph{Attention Maps}
We provide additional visualizations of attention maps on our individual components in Fig.~\ref{fig:supple_fig_attention_map}, and ours compared to other RIS methods~\cite{late_3_cris, joint_1_etris} in Fig.~\ref{fig:supple_fig_attention_map_sotas}.




\section*{Detailed Process of Visualization.} 
\label{appendix:detail_attn}
We also provide the detailed process of attention map visualizations in this section. 
To generate the attention map, we first extract the attention weights from a final layer of the decoder (\eg, a shared transformer layer in ours, and a transformer decoder layer in CRIS~\cite{late_3_cris} and ETRIS~\cite{joint_1_etris}).
We then obtain the attention weight values corresponding to the visual patch tokens from a row of a textual $[\text{CLS}]$ token, since our method and other compared methods predict the segmentation mask based on a similarity map between visual patches and a textual $[\text{CLS}]$ token.
To visualize attention maps on each component of our Shared-RIS, we utilize the attention weights in the final layer in our shared encoder, the average of attention weights of four multi-head self-attentions ($K=4$) for our shared FPN, and the last layer in our shared decoder.

\vfill

\end{document}